\documentclass[runningheads]{llncs}
\usepackage{graphicx}
\usepackage[caption=false, font=footnotesize]{subfig}
\usepackage{multirow}
\usepackage{amssymb}
\usepackage{wrapfig}
\usepackage{caption}
\usepackage{floatflt}
\usepackage{amsmath}
\usepackage{xspace}
\usepackage{appendix}
\usepackage{hyperref}
\newcommand{\myvector}[1]{\mathbf{\lowercase{#1}}}
\newcommand{\set}[1]{\mathcal{#1}}
\newcommand{\realdim}[1]{\mathbb{#1}}
\newcommand{\mymatrix}[1]{\mathbf{\uppercase{#1}}}

\def\model{SCorP\xspace}
\begin{document}

\title{\model: Statistics-Informed Dense Correspondence Prediction Directly from Unsegmented Medical Images}
\titlerunning{\model}

\author{Krithika Iyer\inst{1,2 [0000-0003-2295-8618]} \and 
Jadie Adams\inst{1,2 [0000-0001-7774-5148]} \and
Shireen Y. Elhabian \inst{1,2 [0000-0002-7394-557X]}}

\authorrunning{Iyer et al.}
%
\institute{Scientific Computing and Imaging Institute, University of Utah, UT, USA \and
Kahlert School of Computing, University of Utah, UT, USA \\
\email{krithika.iyer@utah.edu } \email{\{jadie,shireen\}@sci.utah.edu}}

\maketitle

\begin{abstract}
Statistical shape modeling (SSM) is a powerful computational framework for quantifying and analyzing the geometric variability of anatomical structures, facilitating advancements in medical research, diagnostics, and treatment planning. Traditional methods for shape modeling from imaging data demand significant manual and computational resources. Additionally, these methods necessitate repeating the entire modeling pipeline to derive shape descriptors (e.g., surface-based point correspondences) for new data. While deep learning approaches have shown promise in streamlining the construction of SSMs on new data, they still rely on traditional techniques to supervise the training of the deep networks. Moreover, the predominant linearity assumption of traditional approaches restricts their efficacy, a limitation also inherited by deep learning models trained using optimized/established correspondences. Consequently, representing complex anatomies becomes challenging. 
To address these limitations, we introduce \model, a novel framework capable of predicting surface-based correspondences directly from unsegmented images.
By leveraging the shape prior learned directly from surface meshes in an unsupervised manner, the proposed model eliminates the need for an optimized shape model for training supervision. The strong shape prior acts as a teacher and regularizes the feature learning of the student network to guide it in learning image-based features that are predictive of surface correspondences.  
The proposed model streamlines the training and inference phases by removing the supervision for the correspondence prediction task while alleviating the linearity assumption. Experiments on the LGE MRI left atrium dataset and Abdomen CT-1K liver datasets demonstrate that the proposed technique enhances the accuracy and robustness of image-driven SSM, providing a compelling alternative to current fully supervised methods.
\end{abstract}

\section{Introduction}
\vspace{-2mm}
Statistical shape modeling (SSM) is a computational approach for statistically representing anatomies in the context of a population. SSM finds diverse applications in biomedical research, from visualizing organs \cite{borotikar2023statistical}, bones \cite{tufegdzic2022building}, and tumors \cite{mori2023principal}, to assisting in surgical planning \cite{riordan2023modeling}, disease monitoring \cite{zhu2023clinical}, and implant design \cite{friedrich2023point}. Shapes can be represented \textit{explicitly} by a set of ordered landmarks or \textit{correspondence} points, aka point distribution models (PDMs), or implicitly using techniques such as deformation fields \cite{durrleman2014morphometry} or level sets \cite{samson2000level}. This paper focuses on explicit shape representations (i.e., PDMs), characterized by a dense set of correspondences describing anatomically equivalent points across samples. PDM is preferred for its simplicity and efficacy in facilitating interpretable shape comparisons and statistical analyses across populations \cite{cerrolaza2019computational}.

State-of-the-art SSM methods typically require a labor-intensive and computationally demanding workflow that includes manual segmentation of anatomical structures, requiring specialized expertise. Segmentation is followed by pre-processing (e.g., resampling, cropping, and shape registration) and correspondence optimization. This entire process has to be repeated at inference (i.e., for new images), hindering feasibility as an on-demand diagnostic tool in clinical settings.
Deep learning models have emerged as alternatives to traditional tools. Models such as DeepSSM and TL-DeepSSM \cite{bhalodia2018deepssm,bhalodia2024deepssm} learn to estimate correspondences from unsegmented CT/MRI images. Despite their potential, these deep learning approaches still rely on supervised losses and necessitate established PDMs from traditional methods for training. This dependence on established PDMs introduces linearity assumptions, affecting the ability of the models to represent complex anatomical structures adequately. Additionally, this burdensome training requirement inhibits the models' scalability and generalization. 
Furthermore, such deep-learning models depend on shape-based generative data augmentation strategies (via principal component analysis (PCA), non-parametric kernel density estimation (KDE), or Gaussian mixture models), requiring extensive offline computation and imposing time burden. 

\begin{figure}[!ht]
    \vspace{-8mm}
    \centering
    \subfloat[Comparison of requirements and limitations for generation PDMs with various methods\label{fig:subim1}]{%
  \includegraphics[scale=0.75]{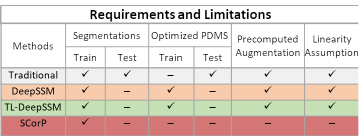}%
}\hfil
\subfloat[Illustration of how the proposed \model and other methods such as  DeepSSM and TL-DeepSSM \cite{bhalodia2018deepssm,bhalodia2024deepssm} are trained for predicting the PDM for shape analysis directly from images.\label{fig:subim2}]{%
  \includegraphics[scale=0.175]{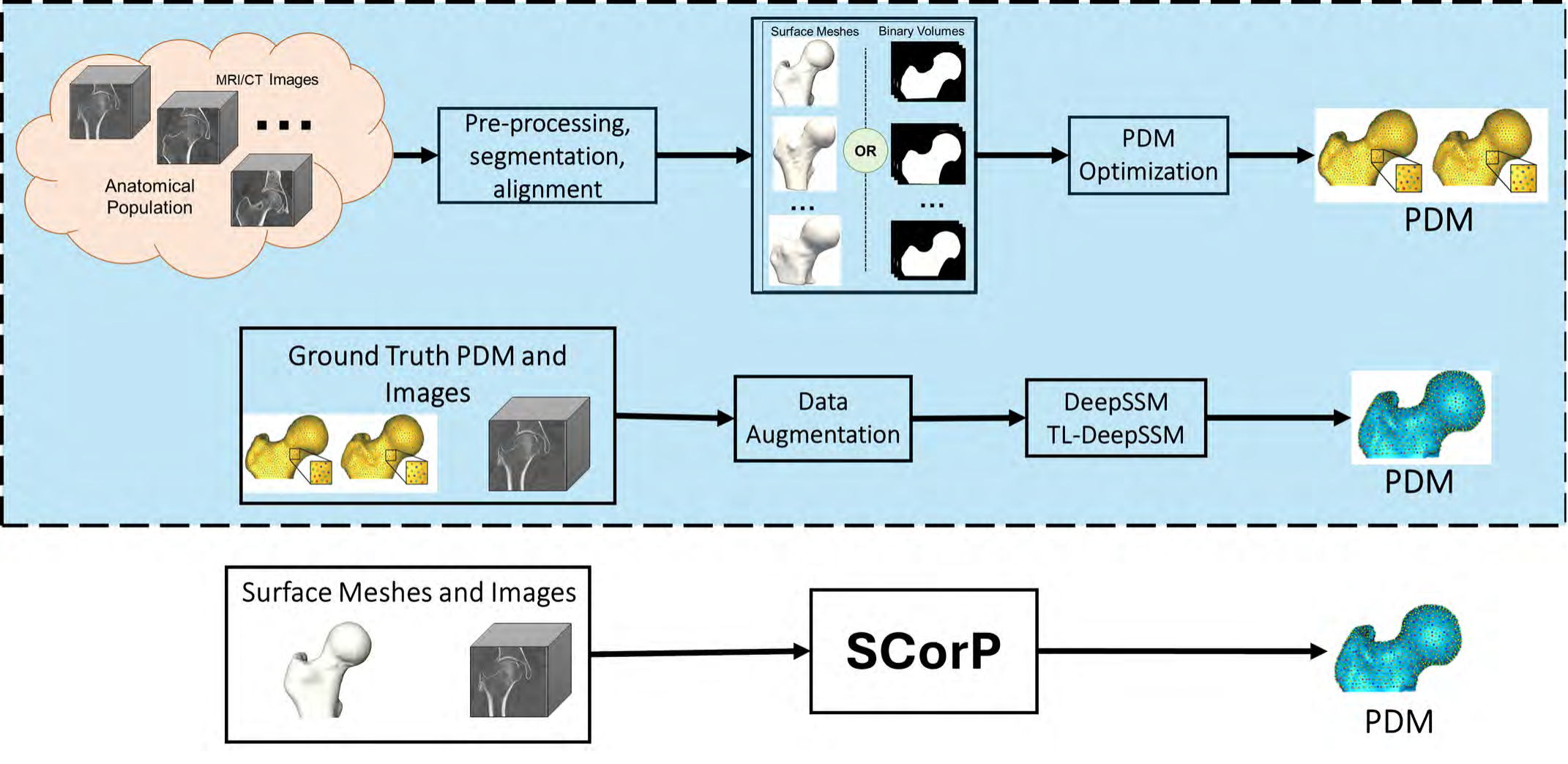}%
}

\caption{\textbf{Comparison of requirements and training pipelines}}
\label{fig:requirement_and_illustration}
\end{figure}
\vspace{-5mm}

The newest breed of SSM deep learning models remove linearity assumptions and drop the requirement for the ground truth PDMs for training \cite{adams2023point2ssm,iyer2023mesh2ssm}. Despite these efforts to improve SSM methodologies, using images directly to predict the correspondences remains challenging. The inherent challenge lies in achieving high-quality shape correspondences from unsegmented images, and supervising the training of these models using an established PDM remains a bottleneck. To tackle these challenges, we propose a novel deep-learning model \model that is capable of predicting correspondences directly from images by leveraging shape prior built directly from the surface representation of anatomies. 
The shape prior can be learned from different surface representations encompassing various forms such as meshes, point clouds, and segmentations, thereby enhancing the model's versatility and applicability.

Volumetric images (e.g., CT/MRI) may present challenges, including (a) noisy and unreliable image features like intensity and texture and (b) poorly defined anatomy boundaries, particularly in low-contrast environments. Furthermore, images depicting irregular shapes with high variability can impede the identification of invariant features. However, when specific anatomical classes are anticipated, integrating shape prior information can guide and constrain the correspondence estimation process to overcome these challenges.
Our proposed model \model takes advantage of existing multi-view data, consisting of paired volumetric images and surface representations, through a \textit{teacher and student} framework. In this framework, shape prior serves as the \textit{teacher} for image-based learning. By guiding the \textit{student} network responsible for feature extraction in the image-driven SSM task, enhancing accuracy and robustness. 

Figure~\ref{fig:requirement_and_illustration} a and b provide an overview of the requirements, limitations, and visual comparison of different SSM pipelines. 
Notably, our proposed method \model distinguishes itself by its minimal requirements (Figure~\ref{fig:requirement_and_illustration}.a), relying solely on surface representation in the form of meshes, point clouds, or binary volumes for training while avoiding adherence to the linearity assumption. Figure~\ref{fig:requirement_and_illustration}.b further provides a visual comparison of the training pipelines of various methods in contrast to \model.
Our main contributions are: 
\begin{enumerate}
    \item We introduce \underline{\textbf{S}}tatistics-informed \underline{\textbf{Cor}}respondence \underline{\textbf{P}}rediction (\textbf{SCorP}), a novel deep learning model designed to predict shape correspondences directly from images. By leveraging the statistics learned from surface representations as a shape prior, our model enables accurate inference of shape descriptors directly from images, bypassing the need for optimization and parameter tuning required in traditional methods.
    \item We validate the accuracy of \model through experiments conducted on the CT (AbdomenCT-1K liver) dataset \cite{Ma-2021-AbdomenCT-1K} and LGE MRI (left atrium) dataset. Furthermore, experiments involving varying training dataset sizes provide evidence of the model's robustness and generalization capabilities.
\end{enumerate}

\section{Related Work}
\vspace{-2mm}
Various traditional methods for establishing correspondences have been proposed, including non-optimized landmark estimation through warping an annotated reference using registration \cite{heitz2005statistical}, parametric methods using basis functions \cite{styner2006framework}, and non-parametric optimization techniques (e.g., particle-based optimization \cite{cates2017shapeworks} and minimum description length (MDL) \cite{davies2002learning}). Non-optimized and parametric methods fail to handle complex shapes due to their fixed geometric basis or predefined template. Non-parametric optimization methods offer a more robust approach by considering the variability of the entire cohort during optimization but still rely on limiting assumptions to define optimization objective (i.e., linearity).

Deep learning models such as DeepSSM and TL-DeepSSM \cite{bhalodia2018deepssm,bhalodia2024deepssm} provide alternatives to traditional SSM tools and are gaining traction. These models learn a functional mapping parameterized by a deep network that estimates surface correspondences from unsegmented images in a supervised manner.
Several models have been proposed to enhance the performance of DeepSSM \cite{bhalodia2018deepssm,bhalodia2024deepssm}. These modifications include incorporating multi-scale and progressive learning modules (e.g., Progressive DeepSSM \cite{aziz2023progressive}), introducing anatomy localization modules for raw images (e.g., LocalizedSSM \cite{ukey2023localization}), and introducing uncertainty quantification (e.g., Uncertain DeepSSM \cite{adams2020uncertain}, VIB-DeepSSM \cite{adams2022images}, BVIB-DeepSSM \cite{adams2023fully}). Despite these advancements, these models still rely on optimized PDMs for training. 
Other deep learning-based image-driven SSM methods have been introduced that leverage radial basis functions (RBF)-based representation to learn control points and normals for surface estimation \cite{xu2023image2ssm}. However, these models face scalability challenges with large datasets and increased correspondences required to model complex anatomies. 

Among the new breed of SSM techniques, Point2SSM \cite{adams2023point2ssm} learns correspondences from unstructured point clouds without connectivity information that represents the surface of the anatomy. However, connectivity information can provide valuable insights when dealing with complex anatomical structures, which leads us to the models that operate on the surface meshes. Models such as FlowSSM \cite{ludke2022landmark} and ShapeFlow \cite{jiang2020shapeflow} employ neural networks to parameterize deformation fields on surface meshes in a low-dimensional latent space, adopting an encoder-free configuration. However, these methods necessitate re-optimization for latent representations of individual mesh samples, posing a notable challenge. Mesh2SSM \cite{iyer2023mesh2ssm} overcomes this issue by replacing the encoder-free setup with geodesic features and EdgeConv \cite{wang2019dynamic} based mesh autoencoder. 

In summary, this review of methods paves the way for our proposed framework, which enhances the image-driven SSM task by directly predicting correspondences from images. Our framework incorporates a principled shape prior and eliminates the need for established PDM supervision during training.
\section{Method}
\vspace{-2mm}
This section presents the formulation, training, and inference phases of \model. Comprehensive details on network architectures and implementation specifics are provided in the Appendix. Surface meshes, point clouds, and binary volumes are all viable forms of surface representation. Without loss of generality, we primarily focus on surface meshes for notation simplicity. However, any surface representation can be used by simply using the relevant architecture for the surface encoder.

Consider a training dataset consisting of \(N\) aligned surface meshes denoted as \(\set{S} = \{S_1,S_2,...S_N\}\) along with their corresponding aligned volumetric images denoted as \(\set{I} = \{\mymatrix{I}_1, \mymatrix{I}_2, ... \mymatrix{I}_N\}\). Each surface mesh is denoted by \(S_j = (\set{V}_j, \set{E}_j)\), where \(\set{V}_j\) and \(\set{E}_j\) denote the vertices and edge connectivity, respectively. The primary objective of the model is to establish a shape prior i.e., \textit{teacher},  by learning to predict a set of \(M\) correspondence points \(\set{C}_j^{S} = \{\mathbf{c}_{j(1)}, \mathbf{c}_{j(2)}, ....\mathbf{c}_{j(M)}\} \) with \( \mathbf{c}_{j(m)} \in \realdim{R}^3\) that comprehensively describe the anatomy represented by the surface mesh \(S_j\). Subsequently, the model leverages this shape prior to guide the feature learning of the image encoder, i.e., \textit{student}, towards extracting image features more conducive to predicting a set of correspondence \(\set{C}_j^{I}= \{\mathbf{c}_{j(1)}, \mathbf{c}_{j(2)}, ....\mathbf{c}_{j(M)}\}\) with \(\mathbf{c}_{j(m)} \in \realdim{R}^3\) directly from the associated image \(\mymatrix{I}_j\). 
\begin{figure}[ht]  
    \centering
    \includegraphics[scale=0.9]{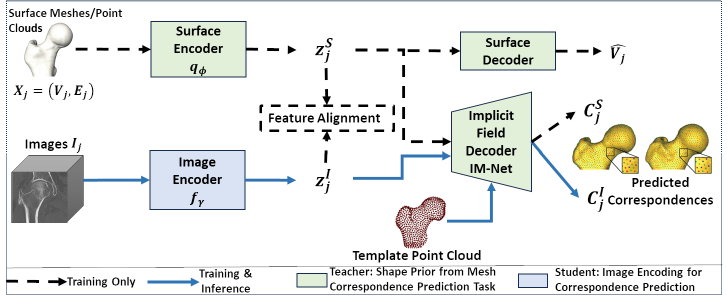}
    \caption{\textbf{Architecture of \model}: Training involves three phases: (1) \textbf{Surface branch training} focuses on shape prior development using the teacher network consisting of the surface autoencoder and IM-NET decoder; (2) \textbf{Image branch embedding alignment} trains the student i.e., image encoder to predict image feature that aligns with the shape prior; (3) \textbf{Image branch prediction refinement} improves predicted correspondences from images.}
    \label{fig:proposed_model}
    \vspace{-8mm}
\end{figure}
\subsection{Surface Autoencoder and Implicit Field Decoder}
To learn the shape prior, i.e., \textit{teacher}, we begin by training a surface autoencoder to learn a low-dimensional representation vector \(\myvector{z}^S_j \in \realdim{R}^L\) for each surface mesh \(S_j = (\set{V}_j, \mymatrix{E}_j)\). We adopt state-of-the-art dynamic graph convolution that employs EdgeConv blocks \cite{wang2019dynamic} (akin to Mesh2SSM \cite{iyer2023mesh2ssm} and Point2SSM \cite{adams2023point2ssm}) to capture permutation invariant local geometric mesh features. EdgeConv blocks compute edge features for each vertex using nearest neighbor computation. These features are then globally aggregated to produce a 1D global descriptor \(\myvector{z}^S_j\) representing the mesh. Notably, the initial EdgeConv block utilizes geodesic distance for feature calculation on the mesh surface. In the case of point cloud data, the original architecture of EdgeConv \cite{wang2019dynamic} graph convolution network without the geodesic information is employed for feature extraction. 

The IM-NET \cite{chen2019net} architecture utilizes the feature vector \(\myvector{z}_j^S\) to predict correspondences \(\set{C}_j^{S}\) for the mesh \(S_j\). This network uses a template point cloud and enforces correspondence relationships across samples by estimating the deformation needed for each point in the template to align with each sample based on \(\myvector{z}_j^S\). IM-NET transforms the template point cloud to match each sample, ensuring consistent correspondence across the dataset.

The surface autoencoder and implicit field decoder are trained jointly to minimize the two-way \(L_2\) Chamfer distance metric between the predicted correspondences \(\set{C}_j^{S}\) and the mesh vertices \(\set{V}_j\) (or point cloud coordinates when considering point clouds), and the reconstruction loss of the autoencoder between the input vertex locations \(\set{V}_j\) and the reconstructed vertex locations \(\hat{\set{V}_j}\). The combined loss function \(\mathcal{L}_{S}\) is expressed as:
\begin{equation}\label{mesh_branch_loss}
    \mathcal{L}_{S} =  \sum_{j=1}^N \left[\mathcal{L}_{CD}(\set{V}_j,\set{C}_j^{S}) +  \alpha \mathcal{L}_{MSE}(\set{V}_j,\hat{\set{V}_j}) \right]
\end{equation}
where \(\alpha\) is the weighting factor for the vertex reconstruction term. 

\subsection{Image Encoder}
The goal of the student network, i.e., the image encoder module, is to learn a compact representation \(\myvector{z}^I_j \in \realdim{R}^L\) for each input image \(\mymatrix{I}_j\). Like surface meshes, the latent representation  \(\myvector{z}^I_j\) will generate the correspondences. To ensure that the encoder captures meaningful representations of the underlying anatomy and is predictive of correspondences, we integrate the shape prior obtained from the teacher, i.e., the surface encoder and implicit field decoder. This integration occurs at two levels: embedding alignment and prediction refinement.
\paragraph{Embedding alignment phase} aligns image features with corresponding surface features, achieved through a regression loss in both surface mesh and image embedding space. Embedding alignment teaches the image encoder to learn representations in the image domain that are semantically meaningful and coherent. Thus, the model learns to map image features to proximal mesh feature regions by minimizing the regression loss in the embedding space. The loss function for image feature alignment is denoted as:
\begin{equation}\label{image_loss_phase2}
\mathcal{L}_{\text{EA}} = \frac{1}{N} \sum_{j=1}^{N} \left[ {| q_{\phi}(\myvector{z}^S_j|S_j) - {f}_{\gamma} (\myvector{z}^I_j|\mymatrix{I}_j)|}^2 \right]
\end{equation}
\paragraph{Prediction refinement phase} refines correspondences predicted by the image branch to match the surface mesh better. Refinement is done by minimizing the Chamfer distance between the predicted correspondences \(\set{C}_j^{I}\) from the image \(I_j\) and the mesh vertices \(\set{V}_j\). This enables the model to refine the initial predictions learned after the embedding alignment phase. The loss function for image branch prediction refinement is denoted as:
\begin{equation}\label{image_loss_phase3}
\mathcal{L}_{\text{PR}} =\sum_{j=1}^N \mathcal{L}_{L_2 CD}(\set{V}_j,{\set{C}_j^{I}})
\end{equation}

\subsection{Training Strategy} \label{training}
\model's training process involves three phases, each focusing on different aspects of the model architecture. The overall loss function guiding the training is formulated as \( \mathcal{L} = \lambda_1 \mathcal{L}_S + \lambda_2 \mathcal{L}_{EA} + \lambda_3 \mathcal{L}_{PR} \) where \(\lambda_1\) and \(\lambda_2\), and \(\lambda_3\) are the weighting factors.
\vspace{-2mm}
\begin{enumerate}
    \item \textbf{Surface branch training}: We begin by training the teacher network consisting of the surface autoencoder and the implicit field decoder. This phase aims to develop a correspondence model based on surface representation, i.e., shape prior. During this phase, the loss function is defined as \(\mathcal{L} = \mathcal{L}_S\), with \(\lambda_1 = 1\) and \(\lambda_2 = \lambda_3 = 0\).
    
    \item \textbf{Image Branch Embedding Alignment}: Next, we focus on training the student network, i.e., the image branch embedding, while keeping the teacher network weights unchanged. This allows the image encoder to learn a shared manifold consistent with the volumetric image and the surface representation. The loss function for this phase is \(\mathcal{L} = \mathcal{L}_{EA}\), with \(\lambda_1 = \lambda_3 = 0\) and \(\lambda_2 = 1\).
    
    \item \textbf{Image Branch Prediction Refinement}: Finally, the predicted correspondences from images are refined to better match the surface meshes while maintaining the feature alignment learned in phase 2 while keeping the teacher network weights unchanged. The loss function for this phase is \(\mathcal{L} = \mathcal{L}_{EA} + \mathcal{L}_{PR}\)  with \(\lambda_1 = 0\) and \(\lambda_2 = \lambda_3  = 1\)
\end{enumerate}
This comprehensive training strategy ensures optimal integration of surface representation based shape prior, for teaching the image encoder to learn representative shape features. During inference on testing samples, correspondences can be directly obtained from an image using the image encoder and the implicit field decoder. Additionally, to enhance the robustness of the surface autoencoder, we introduce vertex denoising as a data augmentation strategy during training. This is achieved by adding jitter to the input vertex positions, encouraging the autoencoder to learn to accurately denoise and reconstruct the original mesh vertices. The same data augmentation strategy can also be extended to point cloud data.




\section{Datasets and Evaluation}
\vspace{-1mm}
\subsection{Datasets}
We select the left atrium and liver datasets for our experiments as they display highly variable shapes, which pose significant challenges for correspondence prediction tasks.\\
\textbf{Left Atrium Dataset (LA): }The dataset comprises 923 anonymized Late Gadolinium Enhancement (LGE) MRIs obtained from distinct patients and were manually segmented by cardiovascular medicine experts. The images were manually segmented at the University of Utah Division of Cardiovascular Medicine, the endocardium wall was used to cut off pulmonary veins.
They have a spatial resolution of \(0.65 \times 0.65 \times 2.5  mm^3\), with the endocardial wall serving as the boundary for the pulmonary veins. Following segmentation, the images were cropped around the region of interest and downsampled by a factor of 0.8 to effectively manage memory usage, resulting in input images of size \(166\times120\times125\). \\
\textbf{AbdomenCT-1K Liver Data:} The dataset \cite{Ma-2021-AbdomenCT-1K} consists of CT scans and segmentations of four abdominal organs, including the liver, kidney, spleen, and pancreas. This dataset comprises 1132 3D CT scans sourced from various public datasets with segmentation verified and refined by experienced radiologists. We use this dataset's CT scans and corresponding liver segmentations for the experiments. The CT scans have resolutions of \(512\times512\) pixels with varying pixel sizes and slice thicknesses between 1.25-5 mm. We visually assess the quality of the images and segmentations and utilize 833 samples. The images were cropped around the region of interest with the help of the segmentations and downsampled by a factor of 3.5 to manage memory usage effectively. The downsampled volume size is \(144 \times 156 \times 115\) with isotropic voxel spacing of 2 mm.

\vspace{-4mm}

\subsection{Models for Comparison}
\vspace{-2mm}
We compare the proposed model against the following:
\vspace{-1mm}
\begin{enumerate}
    \item \textbf{DeepSSM} \cite{bhalodia2024deepssm} is a leading supervised model for predicting correspondence points from 3D image volumes. This method necessitates an optimized PDM for training, where each training instance consists of an image-correspondence pair. We utilize the correspondence supervised version of DeepSSM that uses a fixed decoder initialized with PCA basis and mean shape and trained on mean squared error (MSE) loss between predicted and ground truth correspondences.
   \item \textbf{TL-DeepSSM} \cite{bhalodia2024deepssm} is a variant of DeepSSM designed to overcome limitations associated with PCA usage. However, like DeepSSM, TL-DeepSSM is a supervised approach requiring optimized PDM and image pairs. The TL-variant network architecture \cite{girdhar2016learning} consists of a correspondence autoencoder and a T-flank network for image feature extraction. The network is trained using MSE between predicted and ground truth correspondences for the autoencoder and latent space MSE between the correspondence features and image features. 
   \item \textbf{Baseline} is introduced to demonstrate the effectiveness of introducing shape prior for the image based task. This model consists of an image encoder and an implicit field decoder trained end-to-end to predict correspondences by minimizing the Chamfer distance between the predicted correspondences from images and the mesh vertices.
\end{enumerate}

\subsection{Metrics}
This section describes the metrics used to assess the performance of the quality of the shape models. Since \model does not use the ground truth PDM for training, we exclude the root mean square error metric, which is used in DeepSSM \cite{bhalodia2018deepssm} and TL-DeepSSM \cite{bhalodia2018deepssm,bhalodia2024deepssm}. \\
1. \textbf{Chamfer distance (CD)} measures the average distance from each point in one set (\(C_j\) ) to its nearest neighbor in the other set (\(V_j\)) and vice versa, providing a bidirectional measure of dissimilarity between two point sets.\\
2. \textbf{Point-to-mesh distance (P2M)} is the sum of point-to-mesh face distance and face-to-point distance for the predicted correspondences \(C_j\) and the mesh faces defined using vertices and edges (\(V_j, E_j\)). \\
3. \textbf{Surface-to-surface (S2S) distance} is calculated between the original surface mesh and generated mesh from predicted correspondences. To obtain the reconstructed mesh, we match the correspondences to the mean shape and apply the warp between the points to its mesh.\\
4. \textbf{SSM Metrics:} Three statistical metrics are used to assess SSM correspondence \cite{munsell2008evaluating}. \textbf{Compactness} refers to representing the training data distribution with minimal parameters, measured by the number of PCA modes needed to capture \(95\%\) of variation in correspondence points. \textbf{Generalization} evaluates how well the SSM extrapolates from training to unseen examples, gauged by the reconstruction error (L2) between held-out and training SSM-reconstructed correspondence points. \textbf{Specificity} measures the SSM's ability to generate valid instances of the trained shape class, quantified by the average distance between sampled SSM correspondences and the nearest existing training correspondences. 
\vspace{-4mm}

\subsection{Experimental Setup}
For both datasets, we employ train/test/validation splits of \(80\% / 10\% / 10\% \). We utilize ShapeWorks \cite{cates2017shapeworks}, an open-source shape modeling package, to process images and segmentations (align, crop, binarize segmentations, factor out scale and rotation) and generate surface meshes with 5000 vertices from the segmentations. Additionally, we use ShapeWorks \cite{cates2017shapeworks} to generate the ground truth PDM and follow all prescribed procedures to obtain the required data for training DeepSSM and TL-DeepSSM \cite{bhalodia2024deepssm,bhalodia2018deepssm}. We use the code and hyperparameters provided by the authors of DeepSSM and TL-DeepSSm \cite{bhalodia2024deepssm} to train the models. The PDM is generated with 1024 correspondence particles, sufficient to capture the complex organ shapes. 

We use the medoid shape of each dataset with 1024 correspondences as the template for the implicit field decoder. The medoid shape is identified using the surface-to-surface distances of meshes. This template remains consistent across all three training phases. We employ the Adam optimizer with a fixed learning rate of \(0.00001\) and continue training until convergence, determined through validation evaluation. Convergence is reached when the validation CD does not improve for 200 epochs. The models resulting from the epoch with the best validation CD are chosen for evaluation. Additionally, we set the weighting term of the vertex reconstruction term \(\alpha\) to \(0.001\) (eq.~\ref{mesh_branch_loss}) for all experiments. During the mesh branch training, a random jitter with a standard deviation of \(1\%\) of the maximum vertex size is added as data augmentation. The hyperparameters are identified via tuning for the validation set performance. The source code is available at \href{https://github.com/iyerkrithika21/SCorP_MIUA2024}{https://github.com/iyerkrithika21/SCorP\_MIUA2024}. 

To ensure a fair comparison, we maintain consistency by employing the same architecture for the image encoder and replicating the same image normalization steps used in DeepSSM and TL-DeepSSM \cite{bhalodia2018deepssm,bhalodia2024deepssm} across all experiments. This approach mitigates potential biases arising from architectural variations, facilitating an accurate assessment of performance differences.


\section{Results}
\begin{figure}
    \centering
    \includegraphics[scale=0.90]{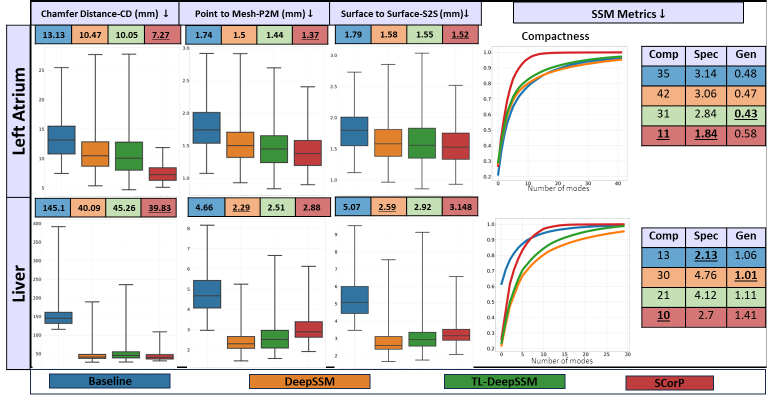}
    \caption{\textbf{Performance metrics} Boxplots illustrating the distribution of performance metrics, with mean values displayed above each plot, for the held-out test samples from the LA and liver datasets. Compactness plots illustrate the cumulative population variation captured by PCA modes, where a larger area under the curve indicates a more compact model. The best metrics are \underline{\textbf{highlighted}} in the figure. Comp = Compactness, Spec = Specificity, Gen = Generalization.}
    \label{fig:metrics_comparison}
    \vspace{-6mm}
\end{figure}

Fig.~\ref{fig:metrics_comparison} shows an overview of the metrics for the held-out test samples of the liver and LA datasets. The baseline method, trained without the mesh-informed shape prior while using the same inference architecture, demonstrates inferior performance for all metrics across the two datasets. This finding emphasizes the critical role of leveraging shape prior information to enhance image-based SSM prediction tasks, especially in the absence of supervision. The proposed model, \model, outperforms the other methods in terms of CD. \model also performs better with respect to P2M and S2S distances for the LA dataset. On the liver dataset, \model exhibits competitive performance with P2M and S2S distances. \model provides the best compactness for both datasets suggesting strong correspondence and showcases comparable specificity and generalization.

\begin{figure}[ht]
    \centering
    \includegraphics[scale=0.33]{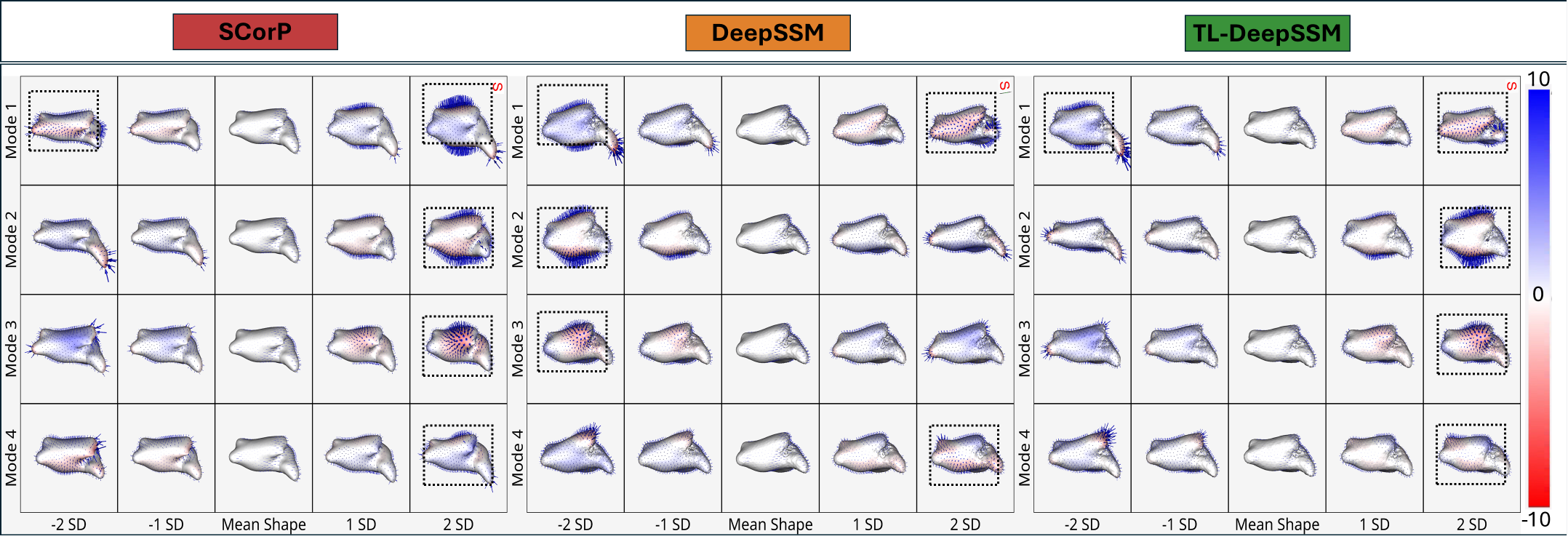}
    \caption{\textbf{PCA modes of variations}: The first four modes of variations of the LA dataset identified by \model, DeepSSM, and TL-DeepSSM \cite{bhalodia2018deepssm,bhalodia2024deepssm}. The color map and arrows show the signed distance and direction from the mean shape. \model shows detailed and smoother variations as compared to the other methods, which are highlighted with boxes.}
    \label{fig:modes_of_variation_la}
\end{figure}

\begin{figure}[ht]
    \centering
    \includegraphics[scale=0.321]{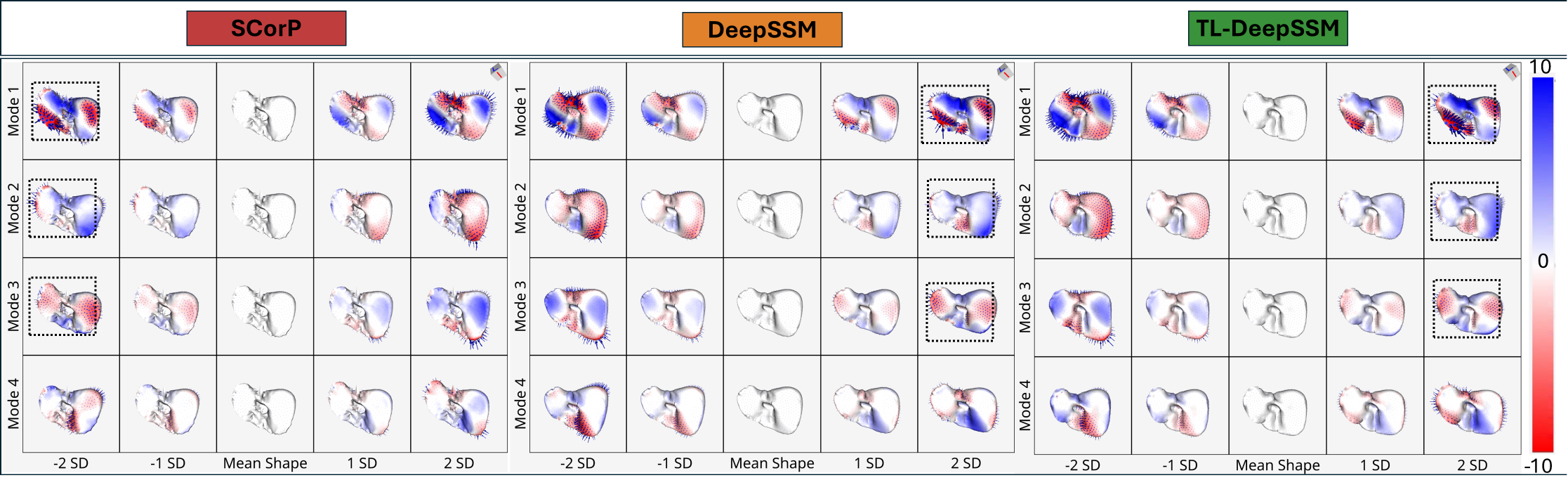}
    \caption{\textbf{PCA modes of variations}: The first four modes of variations of the liver dataset identified by \model, DeepSSM, and TL-DeepSSM \cite{bhalodia2018deepssm,bhalodia2024deepssm}. The color map and arrows show the signed distance and direction from the mean shape. \model shows detailed and smoother variations as compared to the other methods, which are highlighted with boxes.}
    \label{fig:modes_of_variation_liv}
    \vspace{-8mm}
\end{figure}

Furthermore, Fig.~\ref{fig:modes_of_variation_la} and Fig.~\ref{fig:modes_of_variation_liv} depict the top four Principal Component Analysis (PCA) modes of variation identified by \model, DeepSSM, and TL-DeepSSM for the LA and liver datasets, respectively. Despite being an unsupervised method, \model demonstrates competitive performance in identifying modes of variation. Additionally, \model shows detailed and smoother variations as compared to the other methods, which are highlighted with boxes in Fig.~\ref{fig:modes_of_variation_la} and Fig.~\ref{fig:modes_of_variation_liv}. 

\begin{figure}
    \centering
    \includegraphics[scale=0.38]{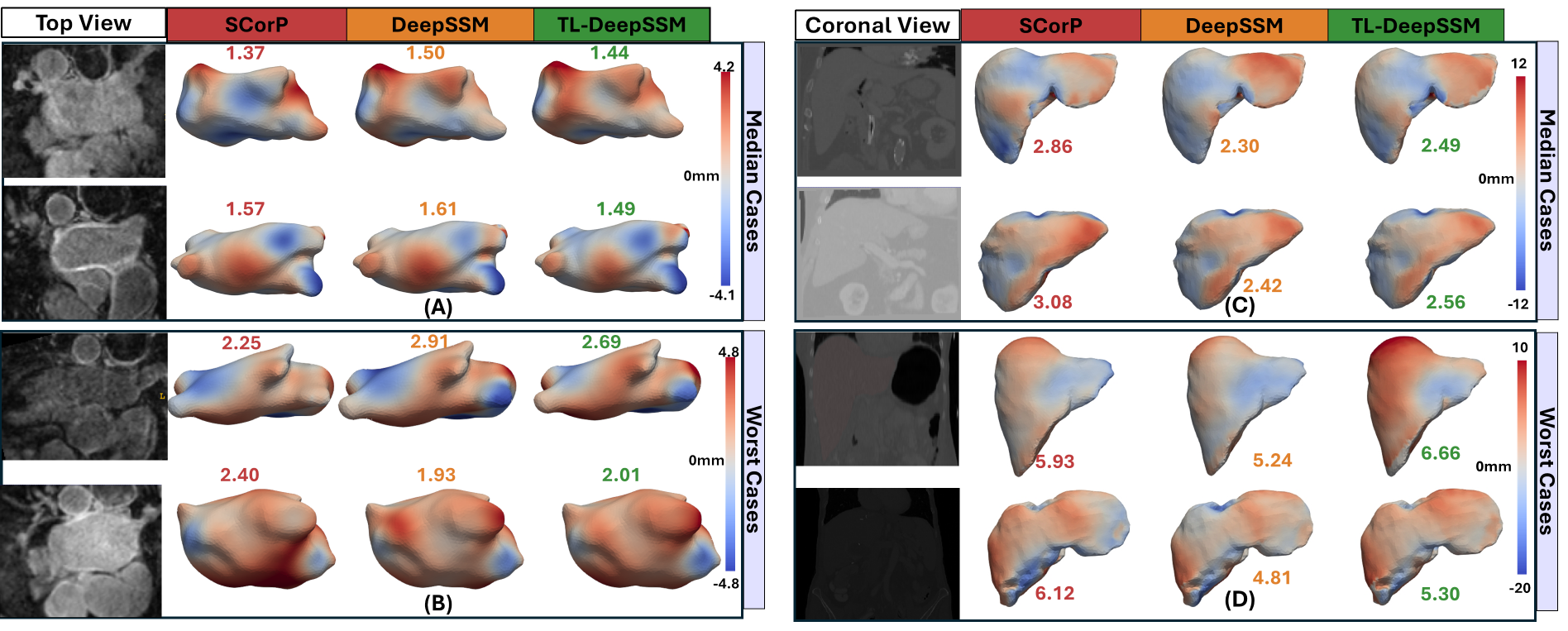}
    \caption{The volumetric image slices of representative samples (worst and median cases) for LA (A and B) and liver (C and D) datasets and all models. The ground truth meshes of the representative samples with a distance map overlay with the correspondence-wise P2M distances for the respective models. The numbers above each sample represent the absolute average P2M distance of the sample. All the models have similar modes of failure and success, and the performances are affected by image quality and the degree of shape outlier.}
    \label{fig:worst_median_p2m}
\end{figure}

We examined the worst and median-performing samples in terms of the P2M distance for all three methods and discovered a substantial overlap among them, suggesting similar success and failure modes. We overlayed the true surface mesh for two median cases and two worst-performing samples with the correspondence-level P2M distances, as depicted in Fig.~\ref{fig:worst_median_p2m}. Additionally, we analyze the corresponding image slices to gain insights into the performance discrepancies. For the  LA dataset, comparing Fig.~\ref{fig:worst_median_p2m}.A and Fig.~\ref{fig:worst_median_p2m}.B reveals the significant impact of image quality on the performance of all three methods. Additionally, a notable observation is the deviation of the shape of the worst-case sample in Fig.~\ref{fig:worst_median_p2m}.B from the population mean (see mean shape in Fig.~\ref{fig:modes_of_variation_la}). Similarly, for the liver dataset, comparing Fig.~\ref{fig:worst_median_p2m}.C and Fig.~\ref{fig:worst_median_p2m}.D highlights clear distinctions in image quality between median and worst-case scenarios. The image slices corresponding to the worst P2M distance exhibit poor contrast and an unclear picture of the liver shape, posing challenges for the image encoder. Notably, when examining Fig.~\ref{fig:worst_median_p2m}, we observe that \model performs comparably with DeepSSM and TL-DeepSSM for median cases. However, for worst-case P2M scenarios in the first row of Fig.~\ref{fig:worst_median_p2m}.C and Fig.~\ref{fig:worst_median_p2m}.D, \model demonstrates superior performance, producing better correspondences for the same samples compared to DeepSSM and TL-DeepSSM. 

\subsection{Ablation Experiments}

\begin{figure}[ht]
    \vspace{-1mm}
    \centering
    \includegraphics[scale=0.95]{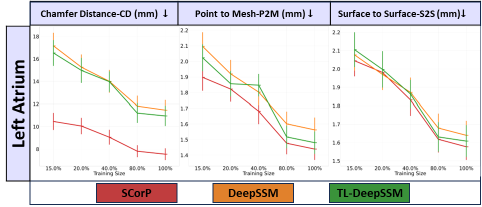}
    \caption{\textbf{Impact of training dataset sizes} The plot illustrates the mean and standard deviation of the performance metrics for all methods at \(15\%, 20\%, 40\%, 80\%, 100\%)\) training dataset size. \model consistency outperforms DeepSSM and TL-DeepSSM and proves to be robust to the dataset size.}
    \label{fig:training_size}
    \vspace{-5mm}
\end{figure}

\paragraph{Impact of training sample size:}We also analyze the robustness of all methods across different training dataset sizes \(15\%, 20\%, 40\%, 80\%, 100\%)\), which yields valuable insights. Fig.~\ref{fig:training_size} illustrates clear trends in mean performance metrics and their standard deviations across various methods at each dataset size. As expected, expanding the training dataset size leads to improved performance across all metrics for all models. Interestingly, even with smaller dataset sizes, \model consistently outperforms DeepSSM and TL-DeepSSM, indicating its robustness and superior generalization ability. One contributing factor to this trend is that the \model does not rely on an optimized PDM during training, unlike DeepSSM and TL-DeepSSM, which depend on the optimized PDM, imposing stronger linearity constraints thereby limiting generalization, particularly with smaller training datasets. \model exhibits greater flexibility and adaptability, allowing it to achieve competitive performance even with limited training data.
\paragraph{Point clouds for surface representation:}To demonstrate the versatility of \model across different surface representation formats, we experimented using the LA dataset, employing point clouds sampled from the meshes to encode the feature vector. In this setup, we utilized the Euclidean distance for k-nearest neighbor calculation in the initial layer of the DGCNN mesh encoder \cite{wang2019dynamic}. Following the training steps outlined in Section~\ref{training}, the model exhibited performance similar to the best-performing model from Fig.~\ref{fig:metrics_comparison} with the following statistics: \textbf{CD \(\mathbf{7.512 \pm 1.72}\), P2M  \(\mathbf{1.435 \pm 0.326}\), and S2S  \(\mathbf{1.56 \pm 0.348}\)}. This highlights that our model is agnostic to the underlying surface representation, enhancing its generalization and usability compared to its counterparts.

\section{Limitations and Future Work}
Given the pivotal role of SSM in diagnostic clinical support systems, it is critical to address the limitations of \model. 
The model currently requires the cohort of images and shapes to be aligned. Relaxing this requirement through developing robust alignment algorithms or exploring alignment-free methods can broaden the usability of \model across various datasets and clinical scenarios.

Furthermore, expanding the framework's capabilities to accommodate diverse data types (sparse slices, orthogonal view slices, and radiography data) and incorporating data augmentation schemes (similar to ADASSM \cite{karanam2023adassm}) holds immense promise for broadening its applicability. Additionally, integrating uncertainty quantification methods to evaluate prediction confidence would enhance result interpretability, further advancing the utility of \model in clinical settings.

\section{Conclusion}
The proposed framework, \model presents a novel approach to inferring correspondences directly from raw images without needing a pre-optimized shape model. By integrating prior shape information from surface representations (meshes, point clouds, binary volumes), \model achieves superior performance compared to traditional and state-of-the-art deep learning methods with less supervision. The three-phase training strategy ensures effective integration of shape statistics-informed priors, guiding the image encoder to learn representative shape features for the correspondence prediction task. Furthermore, \model demonstrates robustness across varying training dataset sizes, highlighting its versatility and applicability in different scenarios. Overall, \model improves upon existing methods by streamlining the PDM generation process, which increases the feasibility of using shape models for research and applications in medical imaging, computer-aided diagnosis, and beyond.

\section{Acknowledgements}
This work was supported by the National Institutes of Health under grant numbers NIBIB-U24EB029011, NIAMS-R01AR076120, and NHLBI-R01HL135568. We thank the University of Utah Division of Cardiovascular Medicine for providing left atrium MRI scans and segmentations from the Atrial Fibrillation projects and the ShapeWorks team.

\bibliographystyle{splncs04.bst}
{\small
\bibliography{paper35_ref}}

\appendix
\section{Appendix}

\subsection{Hyperparamters}
All models were trained on NVIDIA GeForce RTX 2080 Ti. 
\begin{table}[ht]
    \centering
    \begin{tabular}{|c|c|c|}
    \hline
    Parameter & Description & Value\\
    \hline
        B & Batch size & 6  \\
        LR & Learning rate & \(1e^{-5}\)\\
        M & Number of correspondences & 1024 \\
        ES & Early stopping patience epochs & 200 \\
        
    \hline
    \end{tabular}
    \caption{Hyperparameters shared by all models}
    \label{tab:shared_hyperparameters}
\end{table}

\begin{table}[ht]
    \centering
    \begin{tabular}{|c|c|c|}
    \hline
    Parameter & Description & Value\\
    \hline
    L & Latent dimension for \model & 256 \\
        K & Size of neighbourhood for EdgeConv & 20 (LA), 27 (Liver)\\    
        NV & Number of vertices in the mesh & 5000 \\
        
        \hline
    \end{tabular}
    \caption{Hyperparameters for \model}
    \label{tab:hyperparameters}
\end{table}

\subsection{Architecture}
\begin{enumerate}
    \item \textbf{Image encoder:} The encoder architecture utilizes Conv2d layers with \(5\times 5\) filters and the following numbers of filters: \([12, 24, 48, 96, 192]\). After each Conv2d layer, batch normalization and ReLU activation functions are applied. Max pooling layers are incorporated to reduce spatial dimensions. The feature maps are then flattened and passed to the fully connected layers. The fully connected (FC) layer stack consists of linear layers with different input and output feature dimensions: \([193536 -> 384], [384 -> 96], [96 -> 256]\). Each linear layer is followed by a Parametric ReLU (PReLU) activation function. 
    \item \textbf{Surface Autoencoder: } We use the \textit{DGCNN\_semseg\_s3dis} model from the original DGCNN \href{https://github.com/antao97/dgcnn.pytorch/}{Github} repository. 
    \item \textbf{IM-Net:} We use the original implementation of IM-Net from the \href{https://github.com/czq142857/IM-NET-pytorch}{Github} repository. 
\end{enumerate}
\subsection{SSM Metrics}
\begin{enumerate}
    \item Compactness:  We quantify compactness as the number of
PCA modes that are required to capture \(95\%\) of the total variation in the output training cohort correspondence points.
    \item Specificity: We quantify specificity by randomly generating \(J\) samples from the shape space using the eigenvectors and eigenvalues that capture \(95\%\) variability of the training cohort. Specificity is computed as the average squared Euclidean distance between these generated samples and their closest training sample.\\
    \( S= \sum_{\set{C} \in \set{C}_{generated}} ||\set{C} - \set{C}_{train}||^2 \)\\

    \item Generalization: We quantify generalization by assessing the average approximation errors across a set of unseen instances. Generalization is defined as the mean approximation errors between the original unseen shape instance and reconstruction of the shape constructed using the raining cohort PCA eigenvalues and vectors that preserve \(95\%\) variability. \\
    \( G = \sum_{j=1}^U||\set{C}_j - \hat{\set{C}_{j}}||_2^2\) for J unseen shapes.

\end{enumerate}

\end{document}